\pdfoutput=1

\PassOptionsToPackage{table,x11names,dvipsnames}{xcolor}

\documentclass[11pt]{article}

\usepackage[final]{acl}

\usepackage{times}
\usepackage{latexsym}

\usepackage[T1]{fontenc}

\usepackage[utf8]{inputenc}

\usepackage{microtype}

\usepackage{inconsolata}

\usepackage{graphicx}

\usepackage{booktabs}
\usepackage{multirow}
\usepackage[tbtags]{amsmath}
\usepackage{mathtools,amssymb}
\usepackage{enumitem}

\usepackage{array}

\usepackage{algorithm2e}
\usepackage{algpseudocode}

\usepackage[many]{tcolorbox}
\usepackage{setspace}
\usepackage{multicol}
\definecolor{main}{HTML}{F2F2F2}    
\definecolor{sub}{HTML}{CFCFCF}     
\tcbset{
    sharp corners,
    colback = white,
    before skip = 0.3cm,    
    after skip = 0.1cm      
}
\newtcolorbox{boxH}{
    rounded corners,
    arc = 4pt,
    colback = main, 
    colframe = sub, 
    boxrule = 0pt, 
    bottomrule = 2pt,
    rightrule = 1.5pt,
    enhanced,
    left=0.8em, right=0.8em, top=0.4em, bottom=0.4em
}

\usepackage{array}
\renewcommand*{\arraystretch}{1.1}
\usepackage{bbding}

\newcommand\multirowcell[2][0.75]{
\setlength{\tabcolsep}{0pt}
\renewcommand{\arraystretch}{#1}
\begin{tabular}{l}
#2
\end{tabular}
}

\usepackage{caption}
\usepackage{subcaption}
\usepackage{colortbl}

\title{AIPO: Improving Training Objective for Iterative Preference Optimization}

\author{
 \textbf{Yaojie Shen\textsuperscript{1,2*}},
 \textbf{Xinyao Wang\textsuperscript{3*}},
 \textbf{Yulei Niu\textsuperscript{3}},
 \textbf{Ying Zhou\textsuperscript{1,2}},
 \textbf{Lexin Tang\textsuperscript{3}}, 
 \textbf{Libo Zhang\textsuperscript{1,2}},\\
 \textbf{Fan Chen\textsuperscript{3}},
 \textbf{Longyin Wen\textsuperscript{3}},
\\
 \textsuperscript{1}{Institute of Software, Chinese Academy of Sciences, Beijing, China}\\
 \textsuperscript{2}{University of Chinese Academy of Sciences, Beijing, China}\\
 \textsuperscript{3}{ByteDance Inc., San Jose, USA}\\
 \url{https://acherstyx.github.io/project/AIPO}
}

\begin{document}

\maketitle
\let\thefootnote\relax\footnote{{\textsuperscript{*}Equal contribution.}}
\def\thefootnote{\arabic{footnote}} 

\begin{abstract}

Preference Optimization (PO), is gaining popularity as an alternative choice of Proximal Policy Optimization (PPO) for aligning Large Language Models (LLMs).
Recent research on aligning LLMs iteratively with synthetic or partially synthetic data shows promising results in scaling up PO training for both academic settings and proprietary trained models such as Llama3. Despite its success, our study shows that the length exploitation issue present in PO is even more severe in Iterative Preference Optimization (IPO) due to the iterative nature of the process. In this work, we study iterative preference optimization with synthetic data. We share the findings and analysis along the way of building the iterative preference optimization pipeline. More specifically, we discuss the length exploitation issue during iterative preference optimization and propose our training objective for iterative preference optimization, namely \textbf{A}greement-aware \textbf{I}terative \textbf{P}reference \textbf{O}ptimization (AIPO). To demonstrate the effectiveness of our method, we conduct comprehensive experiments and achieve state-of-the-art performance on MT-Bench, AlpacaEval 2.0, and Arena-Hard. Our implementation and model checkpoints will be made available at \url{https://github.com/bytedance/AIPO}.

\end{abstract}

\section{Introduction}

Reinforcement Learning with Human Feedback (RLHF)~\cite{rlhf17,rlhf20} has emerged as a pivotal technique in aligning Large Language Models (LLMs). Although effective relative to Supervised Fine Tuning (SFT)~\cite{instructGPT}, RLHF faces scalability challenges due to the extensive human labeling process required for training data collection. Meanwhile, advancements in both proprietary and open-source LLMs have demonstrated human-level performance across many tasks~\cite{llama3.1, gpt4, qwen2}, suggesting their potential to autonomously generate preference data. Based on this fact, replacing human annotation with LLM-generated data could be a potential solution for the scalability issue aforementioned.\par
Recent studies~\cite{self-rewarding, meta-rewarding, snorkel, spin} support this approach, showing that aligning LLMs with partially generated data in an iterative manner can be effective. However, the length exploitation issue that exists in the generic DPO setting~\cite{r-dpo} is even more amplified in the iterative setting based on our observation and from recent works~\cite{self-rewarding, snorkel}.
In addition, lengthy responses are less efficient for users to consume and cost more hardware resources to generate in practice. For the above reasons, We argue that solely relying on benchmark scores is insufficient to reflect alignment performance and the length exploitation issue needs more research attention. In this work, we showcase our training recipe of aligning LLMs with purely synthetic data iteratively and propose a training objective that is suitable for IPO. Our contributions can be summarized as follows:
\vspace{-0.4em}
\begin{itemize}[leftmargin=*]
\item\textbf{Synthetic Data Curation for Preference Optimization}\hspace{0.5em}
We examine the validity of preference optimization with synthetically generated data. This can be further broken down into instruction creation, response generation, preference ranking, and post-processing. We conclude that models trained with synthetic data yield better performance. $\S$\ref{subsec:data_creation}
\item\textbf{Iterative Training Strategy}\hspace{0.5em}
We define our IPO training strategy and perform ablations on different configurations. During this process, we also observe a more severe length exploitation issue during iterative training with synthetic data. $\S$\ref{subsec:iterative}
\item\textbf{Optimized PO Training Objective}\hspace{0.5em}
We dive deep into the length exploitation issue and discover one of the potential causes is related to the DPO loss. To remedy this, we introduce a new optimized training objective, AIPO, which is more suitable for IPO training scenarios. $\S$\ref{subsec:optimize_dpo}
\end{itemize}

Altogether, we propose an effective training recipe for IPO, including AIPO training objective for IPO training. By leveraging this new training recipe, we achieve state-of-the-art performance on benchmarks including MT-Bench, AlpacaEval 2.0 and Area-Hard.

\section{Preliminaries and Related Work}

\subsection{Direct Preference Optimization}

Direct Preference Optimization (DPO)~\cite{dpo} is derived from the reinforcement learning (RL) phase of the RLHF pipeline~\cite{lamda,stiennon2020learning,bai2022training,ouyang2022training}. The objective of the RL phase is as follows:
\begin{multline}
    \max _{\pi_\theta} \mathbb{E}_{x \sim \mathcal{D}, y \sim \pi_\theta(y \mid x)}\left[r_\phi(x, y)\right]-\\
    \beta \mathbb{D}_{\mathrm{KL}}\left[\pi_\theta(y \mid x) \| \pi_{\mathrm{ref}}(y \mid x)\right],
\end{multline}
where $\pi_{\theta}$ is the policy model, $\pi_{\text{ref}}$ is the reference policy, $r_{\phi}$ is the reward model, and $\beta$ is a hyperparameter to control the deviation from the reference policy. 
Instead of training an explicit reward model and employing RL, DPO reparameterizes the reward utilizing an implicit optimal reward function:
\begin{equation}    \label{eq:reward_func}
    r(x,y) = \beta \log \frac{\pi_{\theta}(y| x)}{\pi_{\mathrm{ref}}{(y| x)}} + \beta \log Z(x),
\end{equation}
where $Z(x)$ is the partition function. 
Incorporating the reward function into the Bradley-Terry model~\cite{bradley1952rank}, 
\begin{equation}    \label{eq:bradley_terry}
    p(y_w \succ y_l | x) = \sigma (r(x,y_w) - r(x,y_l)),
\end{equation}
cancels the partition function and yield the DPO training objective:
\begin{align}   \label{eq:dpo}
    \mathcal{L}_{\mathrm{DPO}}\left(\pi_\theta;\pi_{\mathrm{ref}}\right)=& \notag\\
    -\mathbb{E}_{\left(x, y_w, y_l\right)\sim \mathcal{D}}
    \biggl[\log \sigma\biggl(  \beta& \log \frac{\pi_\theta\left(y_w \mid x\right)}{\pi_{\mathrm{ref }}\left(y_w \mid x\right)}- \notag\\
    \beta& \log \frac{\pi_\theta\left(y_l \mid x\right)}{\pi_{\mathrm{ref}}\left(y_l \mid x\right)} \biggr)\biggr]
    ,
\end{align}
where $y_w$ and $y_l$ are the chosen and rejected responses, respectively.
Consequently, DPO training can be directly applied to binarized preference datasets, which include ternary preference pairs $(x, y_w, y_l)$. DPO eliminates the need for explicit reward model and RL during training, making it more suitable for scaling up the RLHF training stage.

\subsection{Iterative Alignment Methods}

Self-Rewarding~\cite{self-rewarding} focuses on the self-involvement of large language models, utilizing the LLM-as-a-Judge mechanism~\cite{mt-bench} to score their own responses, thereby mitigating the performance bottlenecks that can arise from a frozen judge model. It leverages iterative training to generate self-judgement using an up-to-date model.
Based on Self-Rewarding, Iterative RPO~\cite{iterative-rpo} aims to improve the reasoning ability though IPO by utilizing the Chain-of-Thought (CoT)~\cite{cot} reasoning.
Meta-Rewarding~\cite{meta-rewarding} focuses on improving the self-judging ability in self-rewarding by adding a role of meta-judge to judge the model's own judgement.
sDPO~\cite{sDPO} suggests dividing the available preference datasets into multiple subsets and training on each subset iteratively.
Snorkel-Mistral-PairRM-DPO~\cite{snorkel} is trained iteratively, starting from an initial prompt pool sampled from UltraFeedback. The model is prompted with prompts from UltraFeedback to generate several candidate responses, and then uses PairRM~\cite{pairrm} as reward model to rank the responses. Finally, it trains on the top and bottom responses with DPO.
SPPO~\cite{sppo} approximates the Nash equilibrium iteratively by pushing the chosen rewards to be close to 1/2 and the rejected rewards to be -1/2.
Although all these methods involve IPO, they do not explain the differences between non-iterative and iterative training. Furthermore, they lack a detailed analysis of the design choices and properties involved in IPO training pipeline.

\subsection{Preference Optimization Objectives}

Several PO objectives have been developed in addition to DPO. One line of research is studying PO without relying on a reference model~\cite{cringe,orpo,simpo}. Other methods try to add a margin between with chosen and rejected responses~\cite{ipo, slic-margin, click-margin}. R-DPO~\cite{r-dpo} and SimPO~\cite{simpo} also explore length-controlled approaches. RPO~\cite{rpo} proposes a training objective that incorporates a weighted SFT loss as the regularization term.

\section{Experimental Settings}

\noindent\textbf{Base Model}        \hspace{0.5em}
In our experiments, we use Mistral-7B-Instruct-v0.2~\cite{mistral7b} as the base model for investigating synthetic data curation and iterative training in Sec.~\ref{subsec:data_creation} and \ref{subsec:iterative} due to limited computation resources.
We then use Mistral-Nemo-Instruct-2407~\cite{mistral_nemo}, a more advanced LLM, in Sec.~\ref{subsec:optimize_dpo} for developing our training objective to demonstrate the capability of our method. Note that Mistral Nemo is a 12B model, which acts as a drop-in replacement for Mistral 7B with more capable performance.
We also include additional model ablations on larger-scale language models with their corresponding performance in Tab.~\ref{tab:compare_other_models_and_methods}.

\vspace{0.2em}\noindent
\textbf{Training Data}     \hspace{0.5em}
Following previous works~\cite{snorkel,simpo,sppo}, we use UltraFeedback~\cite{ultrafeedback} as the data source for all experiments.
UltraFeedback is a large-scale preference dataset containing approximately $\mathrm{64K}$ prompts from diverse sources. The responses in UltraFeedback are generated by multiple LLMs and annotated by GPT-4 based on four different aspects: instruction-following, truthfulness, honesty, and helpfulness.
There is also a binarized version\footnote{\url{https://huggingface.co/datasets/HuggingFaceH4/ultrafeedback_binarized}}, which is created by selecting the highest score as the chosen response and one of the remaining as the rejected response.

\vspace{0.2em}\noindent
\textbf{Hyperparameters}   \hspace{0.5em}
For both data curation and training, we set the maximum prompt length to $512$ tokens and the maximum response length to $2048$ tokens.
For data curation ablations in Sec.~\ref{subsec:data_creation}, we train on $\mathrm{60K}$ preference pairs sourced from either UltraFeedback or synthetic data, with a batch size of $128$. 
For iterative training in Sec.~\ref{subsec:iterative}~and~\ref{subsec:optimize_dpo}, we set the batch size to $256$.
After performing ablations for iterative training in Sec.~\ref{subsec:iterative}, we chose to train on $\mathrm{20K}$ preference pairs per iteration by default for the iterative training in Sec.~\ref{subsec:optimize_dpo}.
For all experiments, we train for one epoch for each training stage using the AdamW optimizer with a learning rate of $5\mathrm{e}{-7}$.
We apply a cosine learning rate schedule with $10\%$ warmup.

\vspace{0.2em}\noindent
\textbf{Evaluation}        \hspace{0.5em}
We evaluate our model on three benchmarks: MT-Bench~\cite{mt-bench}, AlpacaEval 2.0~\cite{alpaca_eval,alpaca_eval_lc}, and Arena-Hard~\cite{arena-hard}.
These benchmarks prompt GPT-4-Turbo as an automatic annotator to evaluate the quality of responses.
\textbf{MT-Bench} contains 80 questions across 8 categories, rated using score-based grading on a 10-point scale. We report the average scores on MT-Bench rated by GPT-4-Turbo.
\textbf{AlpacaEval 2.0} contains 805 questions and uses GPT-4-Turbo as both baseline model and judge model. It calculates the win rate against the baseline model and includes a length-controlled win rate, which aims to reduce the impact of length gameability of the LLM judge. For AlpacaEval 2.0, we report the standard win rate (WR), the length-controlled win rate (LC), and the average character length (Avg. Len).
\textbf{Arena-Hard} is an improved version of MT-Bench, including 500 high-quality prompts selected from user queries, using GPT-4 (03/14) as the baseline model and using GPT-4-Turbo as the annotator. It also calculates the win rate against the baseline model. For Arena-Hard, we report win rate (WR) and average token number (Avg. Token).

\begin{table*}[!tbp]
    \centering
    \renewcommand{\arraystretch}{0.85}
    \resizebox{\linewidth}{!}{
    \begin{tabular}{p{5.0em}p{9.2em}ccccccc}
        \toprule
        \multicolumn{2}{c}{\textbf{Training Data}} & \multicolumn{1}{c}{\textbf{MT-Bench}} & \multicolumn{3}{c}{\textbf{AlpacaEval 2.0}} & \multicolumn{2}{c}{\textbf{Arena-Hard}} \\
        \cmidrule(lr){1-2} \cmidrule(lr){3-3} \cmidrule(lr){4-6} \cmidrule(lr){7-8}
        \textbf{Instruction} & \textbf{Response} &
        \textbf{\small GPT-4-Turbo} &
        \textbf{\small LC (\%)} & \textbf{\small WR (\%)} & \textbf{\small Avg. Len} &
        \textbf{\small WR (\%)} & \textbf{\small Avg. Token} \\
        \midrule
        $\mathtt{UF}$       & $\mathtt{UF}$                     &
        6.3             & 20.4          & 16.5          & 1664  & 14.4          & 535 \\
        $\mathtt{UF}$       & $\mathtt{PairRM(UF)}$             &
        6.2             & 20.8          & 16.4          & 1623  & 13.6          & 512 \\
        $\mathtt{UF}$       & $\mathtt{PairRM(Gen(UF))}$        &
        \textbf{6.5}    & 23.7          & 25.2          & 2198  & 17.6          & 649 \\
        $\mathtt{SI(UF)}$   & $\mathtt{PairRM(Gen(SI(UF)))}$    &
        6.4             & \textbf{26.0} & \textbf{28.2} & 2130  & \textbf{19.6} & 615 \\
        \bottomrule
    \end{tabular}
    }
    \caption{The ablation of the synthetic preference pairs for DPO training. $\mathtt{UF}$ indicates the instructions and responses from UltraFeedback, $\mathtt{SI(\cdot)}$ indicates generating self instruct based on the inputs, $\mathtt{Gen(\cdot)}$ indicates generating candidate responses with policy model by taking inputs as prompts, and $\mathtt{PairRM}(\cdot)$ indicates ranking responses by PairRM.}
    \label{tab:data_creation}
\end{table*}

\section{Iterative Preference Optimization with Synthetic Data}

In this section, we document our progress in building the state-of-the-art IPO training recipe. We start with the non-iterative baseline, which trains on the existing pairwise preference dataset, and then move on to iterative training with synthetic data step by step. We investigate the length exploitation issue that occurs during the IPO process and argue that the DPO loss could be one of the potential causes. We then propose our version of the training objective for IPO.

\subsection{Synthetic Data Curation}    \label{subsec:data_creation}

The training data for DPO consists of preference pairs $(x, y_w, y_l)$.
Previous works~\cite{self-rewarding, spin, meta-rewarding} have suggested various methods for creating synthetic preference pairs for preference optimization, but there is a lack of detailed comparison.
To study the roles and effects of different components in the data curation pipeline within IPO training, we conducted a thorough analysis of all aspects, including instructions, responses, and preference rankings, from the non-iterative setting to the iterative setting.
Finally, we propose our own training recipe for the subsequent experiments.

\vspace{0.2em}\noindent
\textbf{External Model Responses vs. Self-Generated Responses}     \hspace{0.5em}
Recent works~\cite{snorkel,simpo,sppo,self-rewarding,iterative-rpo,meta-rewarding} suggest leveraging self-generated responses and existing state-of-the-art reward models to build preference pairs, despite the presence of existing responses and rewards in preference datasets.
To explore this difference, we conduct experiments on the UltraFeedback dataset.
We begin with the vanilla DPO training by training for a single epoch on the $\mathrm{60K}$ preference pairs from UltraFeedback Binarized.
We then replace GPT-4 with PairRM as the ranking method for candidate responses to ensure a fair comparison.
Afterwards, we replace the existing responses with self-generated ones, using PairRM to rank them and select the best and worst responses as $y_w$ and $y_l$, following Snorkel-Mistral-PairRM-DPO~\cite{snorkel}.
The results in Tab.~\ref{tab:data_creation} show that replacing GPT-4 annotations with PairRM rankings (rows 1 vs. row 2) is not crucial for performance, while using self-generated responses (row 3) contributes the most to the performance gap.
We also observe that the average response length significantly increases when using self-generated responses compared to externally-generated ones (row 1\&2 vs.\ row 3\&4).

\vspace{0.2em}\noindent
\textbf{Synthetic Instructions}     \hspace{0.5em}
As suggested in Self-Rewarding~\cite{self-rewarding}, we employ self-instruct~\cite{self-instruct} to build a fully synthetic training pipeline, starting from synthetic instructions. 
Subsequently, the same pipeline is used to generate candidate responses and rank them using PairRM, as mentioned above.
As show in Tab.~\ref{tab:data_creation} (row 4), training with fully synthetic instructions generated by self-instruct achieves performance competitive to human instructions in UltraFeedback.

\vspace{0.2em}\noindent
\textbf{Response Ranking and Data Filtering}        \hspace{0.5em}
Some previous works have leveraged LLMs as judges for response ranking to achieve self-involvement in iterative training~\cite{self-rewarding,iterative-rpo,meta-rewarding}. Although this innovative approach is quite appealing, our experiments indicate that the performance of the LLM judge is not as good as that of existing reward models such as PairRM. Additionally, the LLM judge requires prompting the LLM to generate responses to rank them, which is computationally expensive. 
Therefore, for simplicity and efficiency, we chose to use existing reward models like PairRM.
Additionally, we perform data filtering for synthetic data to ensure quality and diversity.
For more experimental results and details on response ranking and data filtering, please refer to the appendix.

\begin{boxH}
\textit{\textbf{Finding 1:}}\hspace{0.2em}Training with synthetic instructions and responses generated by the current model produces the best performance.
\end{boxH}

\begin{figure*}[tbp]
    \centering
    \includegraphics[width=0.99\linewidth]{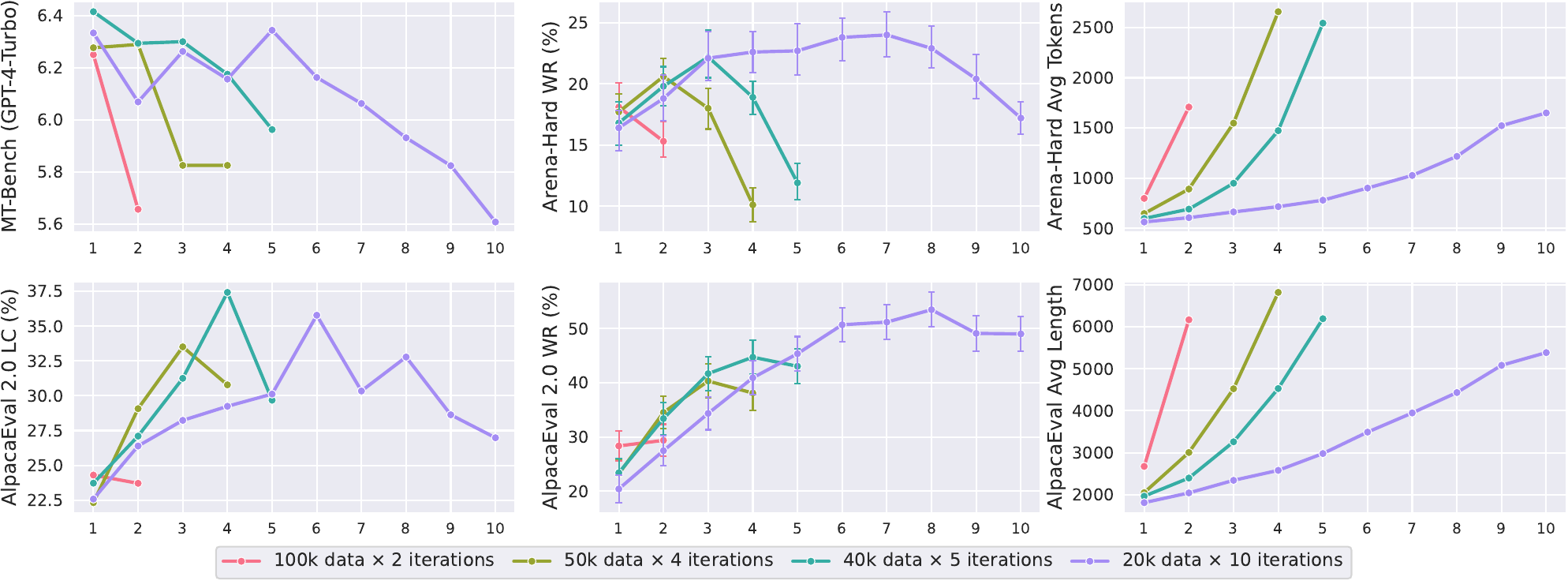}
    \caption{Ablation of iterative training. The horizontal axis represents the training iterations. We train for $T$ iterations, generating $P$ preference pairs in each iteration. In this ablation, we ensure that $T\times P$ remains constant.}
    \label{fig:iterative_performance}
\end{figure*}

\subsection{Iterative Preference Optimization}    \label{subsec:iterative}

Based on our synthetic data curation pipeline, we extend the alignment process to an iterative fashion, as outlined in Algorithm~\ref{alg:iterpo}.
To investigate the impact of iterative training on performance, we keep the total amount of training data constant while varying the data size per iteration.
The results in Fig.~\ref{fig:iterative_performance} show that training over more iterations with a smaller data size per iteration achieves a higher performance upper bound and better training stability, highlighting the importance of iterative training.
We hypothesize that iterative training benefits preference optimization because the data generation model and the reference model for the training objective are updated in each iteration, providing the recent feedback.
However, using a smaller data amount per iteration requires frequent switching of training phases, necessitating a more complex implementation to ensure training efficiency.
Moreover, we found that further reducing the amount of training data for each iteration yields only marginal improvements. 
Therefore, in subsequent experiments, we set our iterative training settings to $\mathrm{20K}$ preference pairs per iteration, with a batch size of $256$ by default.

It is worth noting that compared to the results in Tab.~\ref{tab:data_creation}, although performance is improved, more training steps in iterative training have led to significant length exploitation.
Excessive length exploitation can lead to reduced performance and prevent us from continuously improving performance through iterative training.
To address this issue, we start by analyzing the length exploitation issue in iterative training, and discuss how to alleviate it in Sec.~\ref{subsec:optimize_dpo}.

\begin{boxH}
\textit{\textbf{Finding 2:}}\hspace{0.2em}
Iterative training on synthetic data further improves the performance but amplifies the length exploitation issue.
\end{boxH}

\RestyleAlgo{ruled}
\begin{algorithm}[tbp]
\small
\caption{Iterative Training Pipeline}
\label{alg:iterpo}
\textbf{Input:} $X^{\mathrm{pool}}$: Initial prompt pool, $\theta_0$: Base model, $T$: Number of iterations, $P$: Number of new instructions, $N$: Number of candidate responses. \\
\For{$t=0,\dots,T-1$}{
    \For{$i=1,\dots,P$}{
        Generate new instruction: $x^i = \mathrm{SelfInstruct}_{\theta_{t}}(X^{\mathrm{pool}})$. \\
        \For{$j=1,\dots,N$}{
            Generate candidate responses: $y^i_j \sim p_{\theta_t}(\cdot \mid x^i)$. \\
        }
        Rank responses and obtain preference pairs: $(x^i, y^i_w, y^i_l) = \mathrm{PairRM}(x^i, y^i_1, \dots, y^i_N)$.
    }
    Update model weights: $\theta_{t+1} = \underset{\theta}{\arg\min} \sum_{i=1}^P \mathcal{L}_{\mathrm{AIPO}}(x^i,y^i_w,y^i_l,\theta_{t},\theta)$
}
\textbf{Output:} $\theta_{T}$
\end{algorithm}

\subsection{Revisiting Training Objectives}\label{subsec:optimize_dpo}

\noindent\textbf{Analyzing DPO in Iterative Training}\hspace{0.5em}
Our experiments reveal a significant length exploitation issue in iterative DPO when using self-generated responses, prompting us to investigate the root cause of this undesired behavior.
One advantage of using self-generated responses is that they are generated directly by the model produced in the most recent iteration. They represent the model's best capability in following the provided instructions and both the quality and format of the responses generated for each instruction are very similar. To verify this, we analyze the similarity and log probabilities of the responses from different sources, as shown in Tab.~\ref{tab:response_diff}. Compared to the externally generated responses, the self-generated ones have 1) a much higher average value of the log probabilities for both chosen and rejected responses and 2) a significantly higher similarity between the chosen responses and the rejected responses.
When combined with iterative training, in each iteration, new candidate responses are generated using the latest model weight from the previous iteration, and the reference model is also updated. This process makes the training closer to online RLHF, where the LLM receives real-time AI feedback from the reward model for its own completions.
We hypothesize that this is the primary reason iterative training performs better than non-iterative training, as demonstrated in Sec.~\ref{subsec:iterative}.

However, the similarity of self-generated responses, as indicated in Tab.~\ref{tab:response_diff}, poses a challenge in distinguishing their quality.
Since the candidate responses are similar to each other, it becomes difficult for the reward model to rank them accurately.
Moreover, the training objective in Eq.~\eqref{eq:dpo} aims to increase the log-likelihood of the chosen responses and decrease that of the rejected responses, while ignoring the intrinsic relationship between them.
Forcing the model to distinguish between very similar chosen and rejected responses with high log probabilities can lead to an overestimated gradient value. 
We hypothesize this makes the DPO training objective susceptible to self-generated preference pairs, consequently degrading the model's learning and resulting in responses that are lengthy and less informative.

\begin{figure*}[t]
    \centering
    \includegraphics[width=0.99\linewidth]{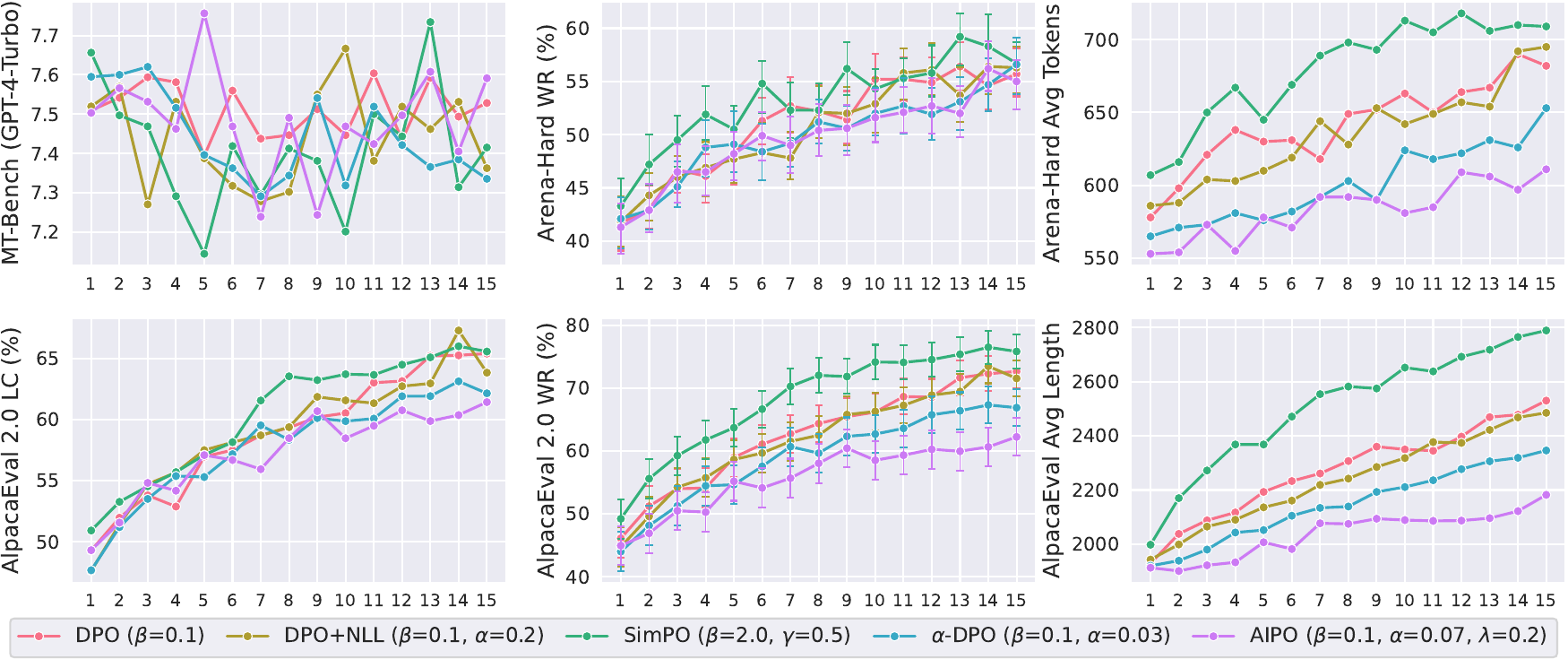}
    \caption{Comparison of different training objectives under iterative alignment setting. The horizontal axis represents the training iterations.}
    \label{fig:training_objective}
\end{figure*}

\vspace{0.2em}
\noindent
\textbf{AIPO: Agreement-Aware Iterative Preference Optimization}      \hspace{0.5em}
We propose to address the difficulty of learning from self-generated responses by leveraging the feedback from the reference model.
To achieve this, we first rewrite the DPO training objective in Eq.~\eqref{eq:dpo} as:
\begin{multline} \label{eq:dpo_new}
    \mathcal{L}_{\mathrm{DPO}} (\pi_\theta ;  \pi_\mathrm{ref}) = \\
    -\mathbb{E}_{\left(x, y_w, y_l\right)\sim \mathcal{D}} 
    \biggl[
    \log\sigma\Bigl(
        \beta \bigl( s_{\theta} -  s_{\mathrm{ref}} \bigr)
    \Bigr)
    \biggr],
\end{multline}
where
\begin{equation*}
\begin{split}
    s_{\theta} &= \log \frac{\pi_\theta\left(y_w \mid x\right)} {\pi_\theta\left(y_l \mid x\right)}
    , \\
    s_{\mathrm{ref}} &= \log \frac{\pi_{\mathrm{ref}}\left(y_w \mid x\right)} {\pi_{\mathrm{ref}}\left(y_l \mid x\right)}
    .
\end{split}
\end{equation*}
We note that $s_{\mathrm{ref}}$ represents the extent to which the reference model model tends to generate the chosen response $y_w$ with a higher probability than the rejected response $y_l$, i.e., the agreement between the reference model and the reward model.
We introduce an additional coefficient $\alpha$ to $s_\mathrm{ref}$ in Eq.~\eqref{eq:dpo_new}. The new training objective is defined as:
\begin{multline}
    \mathcal{L}_{\mathrm{\alpha\text{-}DPO}} (\pi_\theta ;  \pi_\mathrm{ref}) =
    -\mathbb{E}_{\left(x, y_w, y_l\right)\sim \mathcal{D}} \\
    \biggl[
    \log\sigma\Bigl(
        \beta \bigl( s_{\theta} - (1 + \alpha) s_{\mathrm{ref}} \bigr)
    \Bigr)
    \biggr]
    ,
\end{multline}
where we set $\alpha>0$.
We note that $s_\mathrm{ref}$ is not related to the policy model $\pi_\theta$, thus it is equivalent to adding an additional dynamic target reward margin term to the Bradley-Terry model in Eq.~\eqref{eq:bradley_terry}, which can be written as:
\begin{multline*}
p(y_w \succ y_l | x) = \sigma \bigl(r(x,y_w) - r(x,y_l) - \alpha \beta \cdot s_\mathrm{ref}\bigr),
\end{multline*}
where $\alpha \cdot s_{\mathrm{ref}}$ is a dynamic target margin for adjusting the distribution of reward margin. 
\begin{table}[tbp]
    \centering
    \setlength{\tabcolsep}{2pt}
    \resizebox{\linewidth}{!}{
    \begin{tabular}{p{8.5em}ccc}
        \toprule
        \multirow{2}{*}{\textbf{Type of Responses}} &
        \multirow{2}{*}{\multirowcell{\textbf{~Sentence}\\\textbf{Similarity}}} &
        \multicolumn{2}{c}{\textbf{Log Probabilities}} \\
        \cmidrule(lr){3-4}
        & & \textbf{Chosen} & \textbf{Rejected} \\
        \midrule
        \multirowcell{Externally-Generated\\(UltraFeedback)}& 0.64  & -361.5 \small{(-1.140)} & -461.1 \small{(-1.927)} \\[0.6em]
        Self-Generated          & 0.86  & -107.8 \small{(-0.268)} & -110.4 \small{(-0.271)} \\
        \bottomrule
    \end{tabular}
    }
    \caption[]{The similarity and log probabilities of chosen and rejected responses. We use Sentence Transformers\footnotemark~to calculate the similarity, and Mistral-7B-Instruct-v0.2 to generate responses and compute the log probabilities. We also include the length-normalized log probabilities in parentheses.}
    \label{tab:response_diff}
\end{table}\footnotetext{\url{https://sbert.net/}}
\begin{figure}[t]
    \centering
    \includegraphics[width=0.95\linewidth]{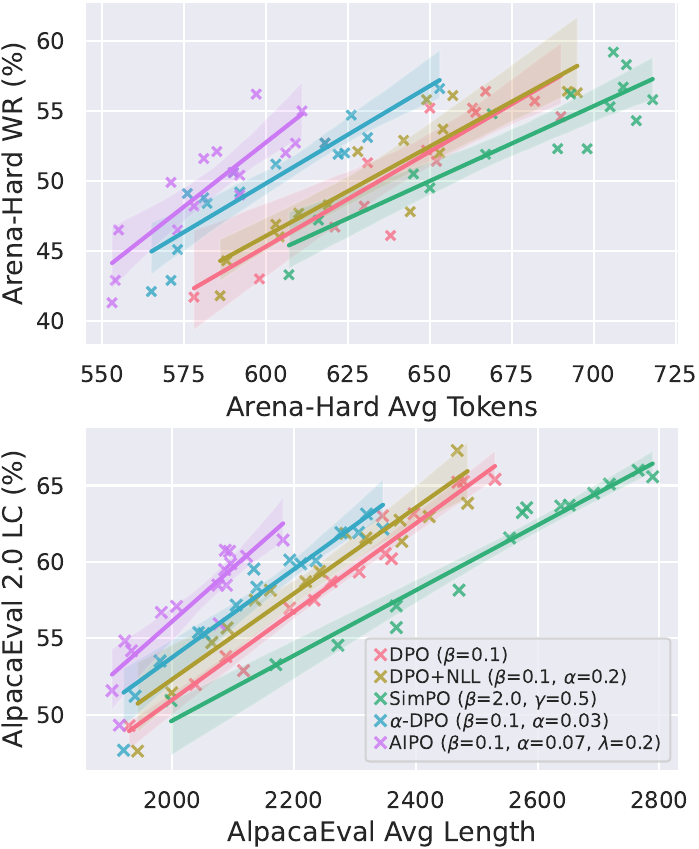}
    \caption{The relationship between the increase in response length and performance with different training objectives. }
    \label{fig:training_objective_by_length}
\end{figure}
As investigated in previous works~\cite{simpo}, a larger target margin value produces a larger reward margin by flatten the reward difference distribution.
Intuitively, $\alpha\beta \cdot s_\mathrm{ref}$ resulting in a larger reward margin when the preference of the reference model agrees with that of the reward model according to the selected chosen and rejected responses, and pose resistance when there is a preference mismatch between the reference model and the reward model.
This ensures a scaling of rewards by considering the agreement between the reference model and the reward model, 
which helps eliminate the aforementioned problem of using self-generated responses.
Next, we analyze the gradient of $\mathcal{L}_{\mathrm{\alpha\text{-}DPO}}$. The gradient with respect to $\theta$ can be written as:
\begin{multline}
    \nabla_\theta \mathcal{L}_{\mathrm{\alpha\text{-}DPO}}\left(\pi_\theta ; \pi_{\mathrm{ref}}\right) = \\ 
    -\beta \mathbb{E}_{\left(x, y_w, y_l\right) \sim \mathcal{D}} 
    \Biggl[
    w_{\theta}
    \cdot
    \biggl(
            \underbrace{
            \nabla_\theta \log \pi\left(y_w \mid x\right)
        }_{\text {increase likelihood of } y_w} - \\
        \underbrace{
            \nabla_\theta \log \pi\left(y_l \mid x\right)
        }_{\text {decrease likelihood of } y_l}
    \biggr)\Biggr],
\end{multline}
where
\begin{equation}    \label{eq:gradient_weight}
\begin{split}
    w_{\theta} &=
    \sigma
    \Bigl(
        \beta \bigl((1+\alpha) \cdot s_\mathrm{ref} - s_{\theta}\bigr)
    \Bigr) \\
    &= 
    \sigma
    \Bigl(
        \beta \bigl(
        \underbrace{s_{\mathrm{ref}} - s_{\theta}}
        _{\substack{\text{weighted by }\\\text{reward estimate}}} + 
        \underbrace{\alpha \cdot s_{\mathrm{ref}}}
        _{\substack{\text{weighted by}\\\text{agreement}}}
        \bigr)
    \Bigr)
\end{split}
\end{equation}
is the gradient weight.
The gradient of the loss function $\mathcal{L}_{\mathrm{\alpha\text{-}DPO}}$ preserves the core properties of DPO: it increases the likelihood of preferred responses $y_w$ and decreases the likelihood of dispreferred responses $y_l$, weighted by the reward estimate $s_{\mathrm{ref}} - s_{\theta}$.
Importantly, $\alpha$-DPO adds an additional weighting term $\alpha\beta \cdot s_\mathrm{ref}$, which weights the gradient by how much more the reference model prefers the chosen response over the rejected response, i.e., the agreement between the preferences of the reference model and the reward model.

\begin{figure}
    \centering
    \includegraphics[width=0.95\linewidth]{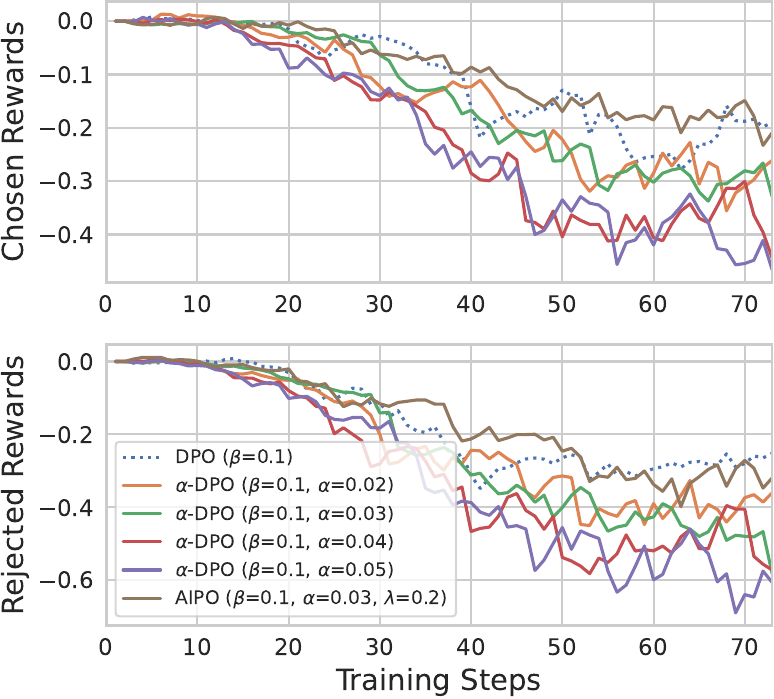}
    \caption{The effect of $\alpha$-DPO on the chosen and rejected rewards during training. }
    \label{fig:alpha_dpo_rewards}
\end{figure}
\begin{table*}[!t]
    \centering
    \setlength{\tabcolsep}{3pt}
    \renewcommand{\arraystretch}{0.95}
    \resizebox{\linewidth}{!}{
    \begin{tabular}{p{23em}cccccc@{}}
        \toprule
        \multirow{2}{*}{\textbf{Methods}} &
        \multicolumn{1}{c}{\textbf{MT-Bench}} &
        \multicolumn{3}{c}{\textbf{AlpacaEval 2.0}} &
        \multicolumn{2}{c}{\textbf{Arena-Hard}} \\
        \cmidrule(lr){2-2} \cmidrule(lr){3-5} \cmidrule(lr){6-7}
        &
        \textbf{\small GPT-4-Turbo} &
        \textbf{\small 2.0 LC (\%)} & \textbf{\small 2.0 WR (\%)} & \textbf{\small Avg. Len} &
        \textbf{\small WR (\%)} & \textbf{\small Avg. Token} \\
        \midrule
        \multicolumn{7}{@{}l}{\textit{Mistral-7B-Instruct-v0.2, UltraFeedback, PairRM}} \\[0.2em]
            \textbf{~~Snorkel-Mistral-PairRM-DPO}~\cite{snorkel}&
            6.2 & 26.4 & 30.2 & 2736 & 20.7 & 564 \\[0.2em]
            \textbf{~~SPPO}~\cite{sppo} &
            6.5 & 28.5 & 31.0 & 2163 & 21.8 & 572 \\[0.2em]
            \rowcolor{SkyBlue2!30!} \textbf{~~AIPO} &
            \textbf{6.7} & \textbf{29.6} & \textbf{37.8} & 2423 & \textbf{22.2} & 663 \\
        \midrule
        \multicolumn{7}{@{}l}{\textit{Llama-3-8B-Instruct, Open Assistant dataset}} \\[0.2em]
            \textbf{~~Self-Rewarding}~\cite{self-rewarding} &
            -   & 34.9 & 34.6 & 1967 & 28.2 & -   \\[0.2em]
            \textbf{~~Meta-Rewarding}~\cite{meta-rewarding} &
            -   & 39.4 & 39.5 & 2003 & 29.1 & -   \\
        \midrule
        \multicolumn{7}{@{}l}{\textit{Llama-3-8B-Instruct, UltraFeedback, PairRM}} \\[0.2em]
            \textbf{~~SPPO}~\cite{sppo} &
            6.8 & 38.8 & 39.9 & 2066 & \textbf{34.0} & 597 \\[0.2em]
            \rowcolor{SkyBlue2!30!} \textbf{~~AIPO} &
            \textbf{6.9} & \textbf{46.2} & \textbf{45.3} & 1977 & 33.8 & 585 \\
        \midrule
        \multicolumn{7}{@{}l}{\textit{Gemma-2-9B-It, UltraFeedback, PairRM}} \\[0.2em]
            \textbf{~~SPPO}~\cite{sppo} &
            7.6 & 53.3 & 47.7 & 1803 & 51.1 & 609 \\[0.2em]
            \rowcolor{SkyBlue2!30!} \textbf{~~AIPO} &
            \textbf{7.7} & \textbf{57.1} & \textbf{54.1} & 1986 & \textbf{56.7} & 701 \\
        \midrule
        \multicolumn{7}{@{}l}{\textit{Others (Ordered by Win Rate on Arena-Hard)}} \\[0.2em]
            \rowcolor{SkyBlue2!30!} \textbf{~~AIPO (Mistral-Large-Instruct-2407)} &
            8.6  & \textbf{67.8} & \textbf{70.8} & 2277 & \textbf{81.6} & 656 \\[0.2em]
            \textbf{~~Claude 3.5 Sonnet (06/20)} &
            -   & 52.4 & 40.6 & 1488 & 79.3 & 567 \\[0.2em]
            \textbf{~~GPT-4 Omni (05/13)} &
            -   & 57.5 & 51.3 & 1873 & 79.2 & 696 \\[0.2em]
            \textbf{~~GPT-4o Mini} &
            -   & 50.7 & 44.7 & 1861 & 74.9 & 668 \\[0.2em]
            \rowcolor{SkyBlue2!30!} \textbf{~~AIPO (Llama-3-70B-Instruct)} &
            8.2 & 60.5 & 60.1 & 2081 & 63.5 & 616 \\[0.2em]
            \rowcolor{SkyBlue2!30!} \textbf{~~AIPO (Gemma-2-27B-It)} &
            8.0 & 57.8 & 48.5 & 1768 & 63.5 & 643 \\[0.2em]
            \textbf{~~Claude 3 Opus (02/29)} &
            -   & 40.5 & 29.1 & 1388 & 60.4 & 541 \\[0.2em]
            \rowcolor{SkyBlue2!30!} \textbf{~~AIPO (Mistral-Nemo-Instruct-2407)} &
            7.4 & 60.4 & 60.7 & 2122 & 56.2 & 597 \\[0.2em]
            \textbf{~~Claude 3 Sonnet (02/29)} &
            -   & 34.9 & 25.6 & 1420 & 46.8 & 552 \\[0.2em]
            \textbf{~~GPT-4 (06/13)} &
            -   & 30.2 & 15.8 & 1140 & 37.9 & 354 \\[0.2em]
            \textbf{~~GPT-3.5 Turbo (06/13)} &
            -   & 22.7 & 14.1 & 1328 & 24.8 & 401 \\[0.2em]
            \textbf{~~Claude 2} &
            -   & 28.2 & 17.2 & 1069 & 24.0 & 295 \\
        \bottomrule
    \end{tabular}
    }
    \caption{The detailed comparison with other IPO methods and proprietary models. For iterative training, we select the model with highest win rate on Arena-Hard.}
    \label{tab:compare_other_models_and_methods}
\end{table*}

In Fig.~\ref{fig:alpha_dpo_rewards}, we investigate the trend of chosen and rejected rewards during training with $\alpha$-DPO.
The results show that $\alpha$-DPO leads to a decrease in both chosen and rejected response log probabilities, which might be harmful as noted by previous works~\cite{iterative-rpo}.
Interestingly, however, the performance of $\alpha$-DPO improves despite the decrease in log probabilities for both the chosen and rejected responses by the policy model.
We highlight that the decrease in probabilities of self-generated responses also indicates the shift of policy model's output distribution, reflecting that $\alpha$-DPO provides a clear target for learning preferences, thereby making it easier for the model to learn preferences.
Since rapid changes in the output distribution may be unstable, we employ Negative Log Likelihood (NLL) loss as a compensatory measure, following previous works~\cite{iterative-rpo,llama3.1}. The NLL term is defined as:
\begin{equation}
    \mathcal{L}_\mathrm{NLL} = - \frac{1}{|y_w|} \log \bigl( \pi_{\theta}(y_w \mid x) \bigr).
\end{equation}
Combining $\alpha$-DPO with NLL term, our AIPO training objective for IPO is defined as:
\begin{equation}
\mathcal{L}_{\mathrm{AIPO}} = \mathcal{L}_{\mathrm{\alpha\text{-}DPO}} + \lambda \cdot \mathcal{L}_{\mathrm{NLL}}
\end{equation}
where $\lambda$ balances the relative importance of $\mathcal{L}_{\mathrm{NLL}}$.

\vspace{0.2em}
\noindent
\textbf{Experiments and Comparisons}    \hspace{0.5em}
We conduct comprehensive experiments to compare different methods in iterative settings. To ensure a fair comparison, we perform hyperparameter searches for each method to find the best combination of hyperparameters. For more details, please refer to the appendix.
As shown in Fig.~\ref{fig:training_objective}, although the performance of SimPO increases rapidly with more training iterations, it produces much longer responses compared to other methods. This contrasts with the goal of length normalization included in the SimPO training objective, indicating that length normalization alone is not suitable for self-generated responses in the IPO training pipeline.
In contrast, $\alpha$-DPO achieves competitive performance with much shorter responses in iterative settings.
The advantage of $\alpha$-DPO becomes more apparent when comparing performance under the same length constraints, as shown in Fig.~\ref{fig:training_objective_by_length}.
Additionally, the results show that adding NLL term to DPO has only a marginal effect on performance, demonstrating that its primary role is to stabilize the distribution of the policy model during training.
However, when the NLL term is added to $\alpha$-DPO, forming AIPO, the trend of reward decline is effectively mitigated, as shown in Fig.~\ref{fig:alpha_dpo_rewards}, and performance is significantly enhanced under the same length constraints, as illustrated in Fig.~\ref{fig:training_objective_by_length}.
This underscores the importance of the NLL term in AIPO for stabilizing the output log-likelihood distributions of the winning and losing pairs of the policy model, ultimately leading to better overall performance.

In Tab.~\ref{tab:compare_other_models_and_methods}, we compare AIPO with other IPO training methods and conduct scaling experiments to verify the scaling ability of AIPO on large LLMs, comparing it with proprietary models. 
Due to the prohibitive training costs, we reuse the best set of hyperparameters from the ablation studies on Mistrial Nemo. 
The results in Tab.~\ref{tab:compare_other_models_and_methods} show that, compared to other methods, AIPO achieves competitive or superior performance without a significant increase in response length. We observe that training AIPO on Mistrial 7B yields less performance improvement than training on Llama 3 8B or Gemma, indicating the necessity of using stronger base models to ensure continuous performance growth in IPO training.

\section{Conclusion}

In this work, we explore the transition from non-iterative DPO training to iteratively aligning LLMs with synthetic data.
Our study addresses the gap in existing works concerning the comparison and analysis of IPO training pipelines, offering a robust recipe for IPO training.
Based on this, we investigate the issue of length exploitation in iterative training, which severely impacts performance. We analyze the properties of self-generated responses and propose AIPO, which aims to achieve agreement-aware adjustments in the training objective to alleviate the length issue and ensure stability in IPO training.
Combing these techniques, we achieve state-of-the-art performance on MT-Bench, AlpacaEval 2.0 and Arena-Hard.

\section*{Limitations}

\begin{enumerate}
\item In our experiments, we use the same set of hyperparameters for each iteration, ignoring the potential differences between iterations. Performing specific parameter adjustments for each iteration could potentially yield different results, which have not been fully explored in this study.
\item In this paper, we primarily focus on mitigating length issues by optimizing the training objective.
However, beyond the training phase, other strategies could also help alleviate length issues.
These include controlling model generation when producing candidate responses, considering length factors when selecting chosen and rejected responses, and thoroughly cleaning the training data, etc.
We leave these as future works.
\end{enumerate}


\bibliography{custom}

\appendix

\section{Appendix}

\subsection{Experiments for Response Ranking and Data Filtering}

Following the experimental settings in Sec.~\ref{subsec:data_creation}, we perform experiments to compare methods for response ranking and data filtering.

\vspace{0.2em}
\noindent\textbf{Response Ranking}      \hspace{0.5em}
Existing works primarily use two methods to rank candidate responses: 1) using existing reward models (such as PairRM) and 2) prompting LLMs with a judge prompt to score responses. The LLM judge model can either be external or the model itself. We perform ablation on these methods,
and the results in Tab.~\ref{tab:ranking_responses} indicate that PairRM outperforms LLM judge across all benchmarks.
For simplicity and optimal performance, we employ PairRM as the judge model for ranking candidate responses.

\vspace{0.2em}
\noindent\textbf{Filtering and Data Cleaning}   \hspace{0.5em}
Our IPO training pipeline relies heavily on synthetic data, which makes it sensitive to the characteristics of the base model, such as response style, diversity, and capacity. For smaller-scale models, the synthetic data often contains more biases and unexpected responses. To address this, we implement basic rule-based filtering and data cleaning strategies during the self-instruct creation and candidate response generation stages, aiming to stabilize the training and mitigate biases. As shown in Tab.~\ref{tab:data_filtering}, applying simple rule-based filtering slightly improves overall performance. We thus apply filtering by default for experiments in main paper. 

\subsection{Model Ablation for Iterative Preference Optimization}

To investigate how the base model affects length issue in IPO training, we perform model ablation in IPO, and employ vanilla DPO in the training phase.
The results are shown in Fig.~\ref{fig:model_ablation}.
Our conclusion is that a stronger base model generally experiences fewer length issues.
At the same time, length exploitation also appears to be influenced by the training methods and training data. 
For instance, both Mistral-7B-Instruct-v0.2 and Mistral-7B-Instruct-v0.3 exhibit severe length problems during IPO training.
The pre-training of LLMs is a broad topic that extends beyond the scope of this work, and many details of the pre-training processes are often opaque. Therefore, we only highlight the potential impact of pre-training based on our observations.

\begin{table}[tbp]
    \centering
    \setlength{\tabcolsep}{3pt}
    \renewcommand{\arraystretch}{0.9}
    \resizebox{\linewidth}{!}{
    \begin{tabular}{p{6em}cccc}
        \toprule
        \multirow{2}{6em}{\textbf{Response Ranking}} &
        \multicolumn{1}{c}{\textbf{MT-Bench}} &
        \multicolumn{2}{c}{\textbf{AlpacaEval 2.0}} &
        \multicolumn{1}{c}{\textbf{Arena-Hard}} \\
        \cmidrule(lr){2-2} \cmidrule(lr){3-4} \cmidrule(lr){5-5}
        &
        \textbf{\small GPT-4-Turbo} &
        \textbf{\small LC (\%)} & \textbf{\small WR (\%)} &
        \textbf{\small WR (\%)} \\
        \midrule
        Self-Reward  & 6.1          & 21.0          & 22.5          & 14.5          \\
        External LLM & 6.1          & 20.5          & 19.6          & 13.9          \\
        PairRM       & \textbf{6.2} & \textbf{23.9} & \textbf{25}   & \textbf{18.8} \\
        \bottomrule
    \end{tabular}
    }
    \caption{Comparison of methods for ranking candidate responses.
    We generate candidate responses for 60K instructions from UltraFeedback. For LLM judge, we use the LLM-as-a-Judge prompt from Self-Rewarding. When using external LLM as the judge, we employ Mixtral-8x7B-Instruct-v0.1 as the judge model.}
    \label{tab:ranking_responses}
\end{table}

\begin{table}[tbp]
    \centering
    \setlength{\tabcolsep}{3pt}
    \renewcommand{\arraystretch}{0.9}
    \resizebox{\linewidth}{!}{
    \begin{tabular}{p{6em}ccccc}
        \toprule
        \multirow{2}{6em}{\textbf{Response Ranking}} &
        \multicolumn{1}{c}{\textbf{MT-Bench}} &
        \multicolumn{2}{c}{\textbf{AlpacaEval 2.0}} &
        \multicolumn{1}{c}{\textbf{Arena-Hard}} \\
        \cmidrule(lr){2-2} \cmidrule(lr){3-4} \cmidrule(lr){5-5}
        &
        \textbf{\small GPT-4-Turbo} &
        \textbf{\small LC (\%)} & \textbf{\small WR (\%)} &
        \textbf{\small WR (\%)} \\
        \midrule
        w/o Filtering   & 6.3           & 25.5          & 27.9          & 19.0          \\
        w/ Filtering    & \textbf{6.4}  & \textbf{26.0} & \textbf{28.2} & \textbf{19.6} \\
        \bottomrule
    \end{tabular}
    }
    \caption{Effect of rule-based filtering on training with synthetic data.}
    \label{tab:data_filtering}
\end{table}

\begin{figure}[!t]
    \centering
    \includegraphics[width=\linewidth]{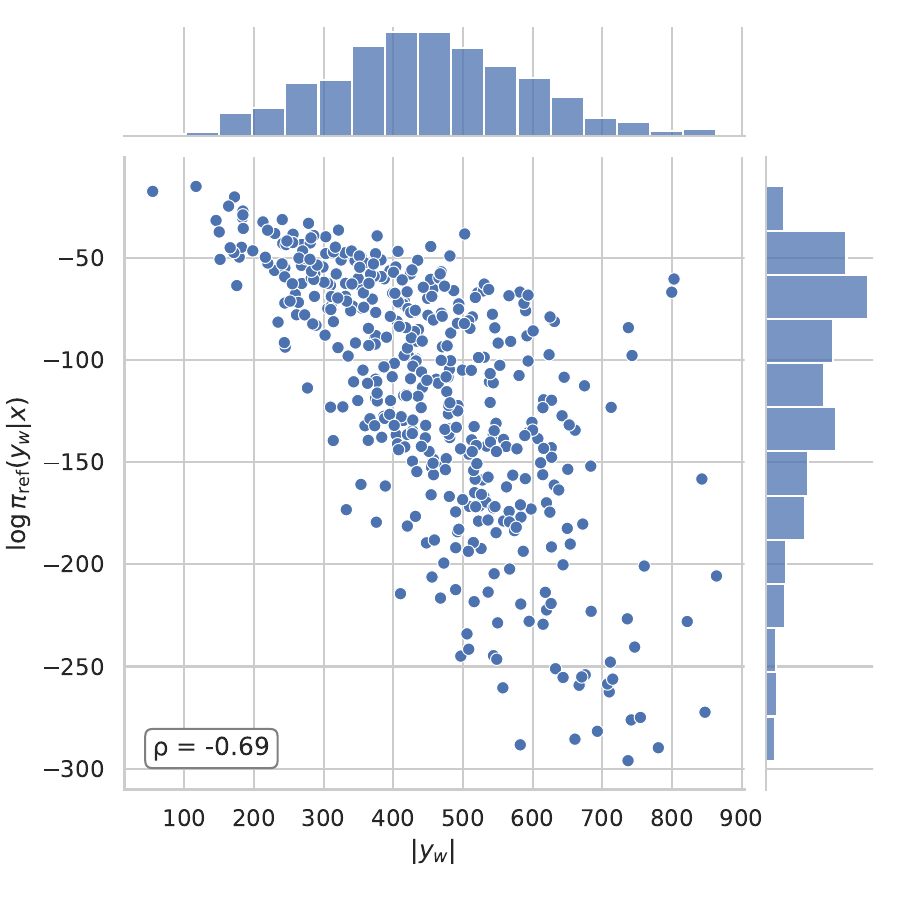}
    \caption{The distribution of response lengths and the corresponding log probabilities by the reference model.}
    \label{fig:length_vs_logps}
\end{figure}

\begin{figure}[!t]
    \centering
    \includegraphics[width=\linewidth]{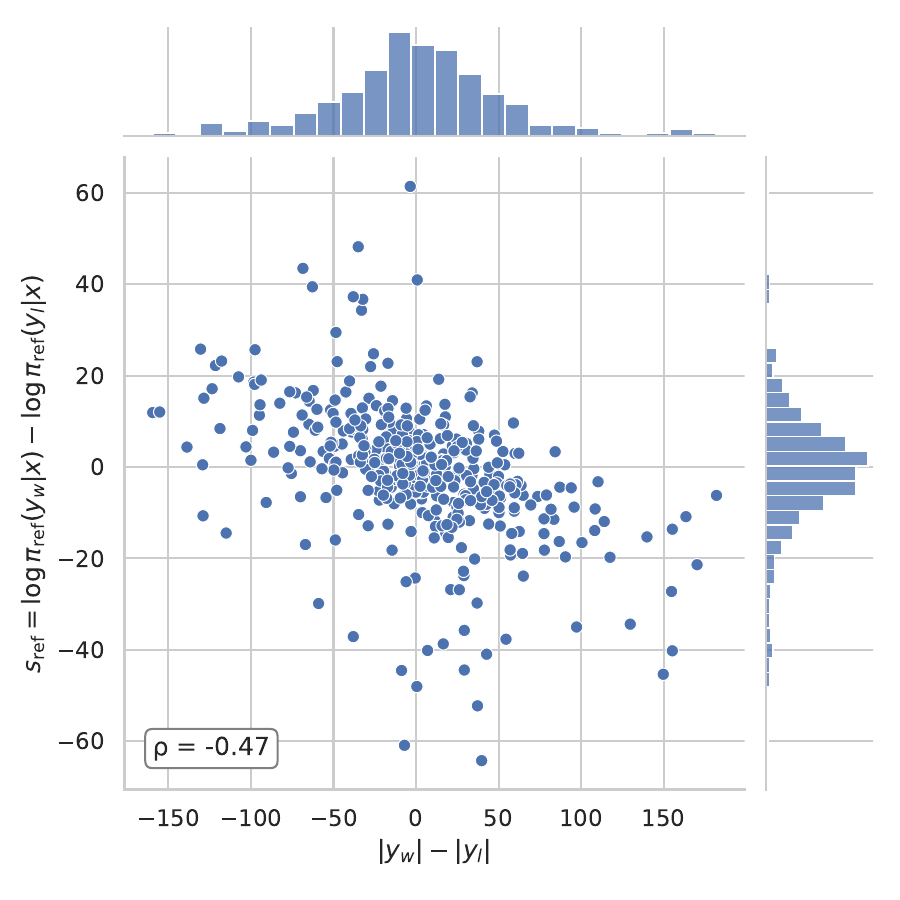}
    \caption{The distribution of $s_{\mathrm{ref}}$ and the difference in length between chosen and rejected responses.}
    \label{fig:s_ref_vs_length_diff}
\end{figure}

\begin{table*}[p]
    \centering
    \renewcommand{\arraystretch}{0.7}
    \resizebox{0.999\linewidth}{!}{
    \begin{tabular}{p{0.22\textwidth}p{0.75\textwidth}p{0.27\textwidth}}
    \toprule
    \textbf{Method} & \textbf{Objective} & \textbf{Hyperparameter} \\
    \midrule
        \multirowcell{\textbf{DPO}\\\cite{dpo}} &
        $-\log \sigma \Bigl(\beta \log \frac{\pi_\theta(y_w \mid x)}{\pi_\mathrm{ref}(y_w \mid x)} - \beta \log \frac{\pi_\theta(y_l \mid x)}{\pi_\mathrm{ref}(y_l \mid x)}\Bigr)$ &
        $\beta \in [0.1, 0.5, 1.0]$ \\
    \midrule
        \multirowcell{\textbf{DPO+NLL}\\\cite{iterative-rpo}} &
        $-\log \sigma \Bigl(\beta \log \frac{\pi_\theta(y_w \mid x)}{\pi_\mathrm{ref}(y_w \mid x)} - \beta \log \frac{\pi_\theta(y_l \mid x)}{\pi_\mathrm{ref}(y_l \mid x)}\Bigr) - \frac{\alpha}{\lvert y_w\rvert} \log \bigl( \pi_\theta (y_w \mid x) \bigr)$ &
        \multirowcell[0.8]{
        $\beta \in [0.1]$ \\
        $\alpha \in [0.2, 0.4, 0.6]$
        } \\
    \midrule
        \multirowcell{\textbf{SimPO}\\\cite{simpo}} &
        $-\log \sigma \Bigl( \frac{\beta}{\lvert y_w\rvert}\log\pi_\theta(y_w\mid x) - \frac{\beta}{\lvert y_l\rvert}\log\pi_\theta(y_l\mid x) - \gamma \Bigr)$ &
        \multirowcell[0.8]{
        $\beta \in [2.0, 3.0]$ \\
        $\gamma \in [0.5, 1.0]$
        } \\
    \midrule
        \textbf{\boldmath{$\alpha$}-DPO} &
        $-\log\sigma \Bigl( \beta \log \frac{\pi_\theta(y_w\mid x)}{\pi_\theta(y_l\mid x)} - (1+\alpha) \beta \log \frac{\pi_\mathrm{ref}(y_w\mid x)}{\pi_\mathrm{ref}(y_l\mid x)} \Bigr)$ &
        \multirowcell[0.8]{
        $\beta \in [0.1]$ \\
        $\alpha \in [0.02, 0.03, 0.04, 0.05]$
        } \\
    \midrule
        \textbf{AIPO} &
        $-\log\sigma \Bigl( \beta \log \frac{\pi_\theta(y_w\mid x)}{\pi_\theta(y_l\mid x)} - (1+\alpha) \beta \log \frac{\pi_\mathrm{ref}(y_w\mid x)}{\pi_\mathrm{ref}(y_l\mid x)} \Bigr) - \frac{\lambda}{\lvert y_w\rvert} \log \bigl( \pi_\theta (y_w \mid x) \bigr)$ &
        \multirowcell[0.8]{
        $\beta \in [0.1]$ \\
        $\alpha \in [0.03, 0.05, 0.07]$ \\
        $\lambda \in [0.2, 0.5]$
        } \\
    \bottomrule
    \end{tabular}
    }
    \caption{The range of hyperparameter search for each training objective.}
    \label{tab:hyperparameter}
\end{table*}

\subsection{Studying the Relation of Agreement-Awareness to Length Regularization}

We perform a statistical analysis of the response lengths and log probabilities, the results are shown in Fig.~\ref{fig:length_vs_logps}. 
The results indicate that response length has a strong correlation with log probabilities: as the response length increases, the log probabilities tend to decrease. 
This is due to the inherent diversity of the model during the generation process.
Examining the gradient weight in Eq.~\eqref{eq:gradient_weight}, we observe that after expressing $s_{\mathrm{ref}}$ as the difference in log probabilities of the reference model for chosen and rejected responses, the gradient weight can be written as:
\begin{multline} \label{eq:gradient_weighting_appendix}
    w_{\theta} =
    \sigma
    \bigl(
        \beta \bigl(
        s_{\mathrm{ref}} - s_{\theta}
        + \\
        \alpha \bigl(
            \log \pi_{\mathrm{ref}}(y_w\mid x) - \log \pi_{\mathrm{ref}}(y_l\mid x) 
            \bigr)
        \bigr)
    \bigr).
\end{multline}
When combined with the relationship between log probabilities and response length, Eq.~\eqref{eq:gradient_weighting_appendix} appears to resemble the length regularization used in R-DPO~\cite{r-dpo}. The additional term $\alpha\beta \cdot s_{\mathrm{ref}}$ gives a higher gradient weight when the chosen response is shorter than the rejected one, and a lower gradient weight when the chosen response is longer.
This suggests the needs of further analyze the relationship between $s_\mathrm{ref}$ and the length difference $|y_w| - |y_l|$.
In the Fig.~\ref{fig:s_ref_vs_length_diff}, we plot the distribution of $s_\mathrm{ref}$ and $|y_w| - |y_l|$. We can see that although there is some correlation between $s_{\mathrm{ref}}$ and the length difference, its value is lower than that presented in Fig.~\ref{fig:length_vs_logps} ($\rho=-0.47$ vs. $\rho=-0.69$).
This is mainly because AIPO takes into account the preferences of the reference model, which makes the AIPO acts differently from length regularization.

\subsection{Hyperparameter Tuning}

\begin{figure*}[tbp]
    \centering
    \begin{subfigure}[t]{0.325\linewidth}
        \includegraphics[width=1\linewidth]{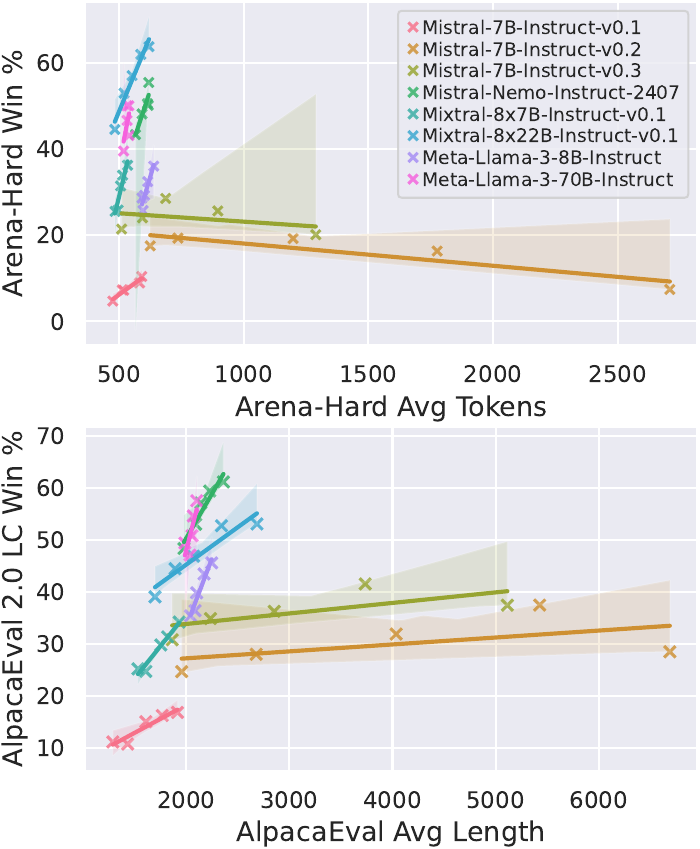}
        \caption{The ablation of base model for iterative preference optimization.}
        \label{fig:model_ablation}
    \end{subfigure}
    \begin{subfigure}[t]{0.325\linewidth}
        \includegraphics[width=\linewidth]{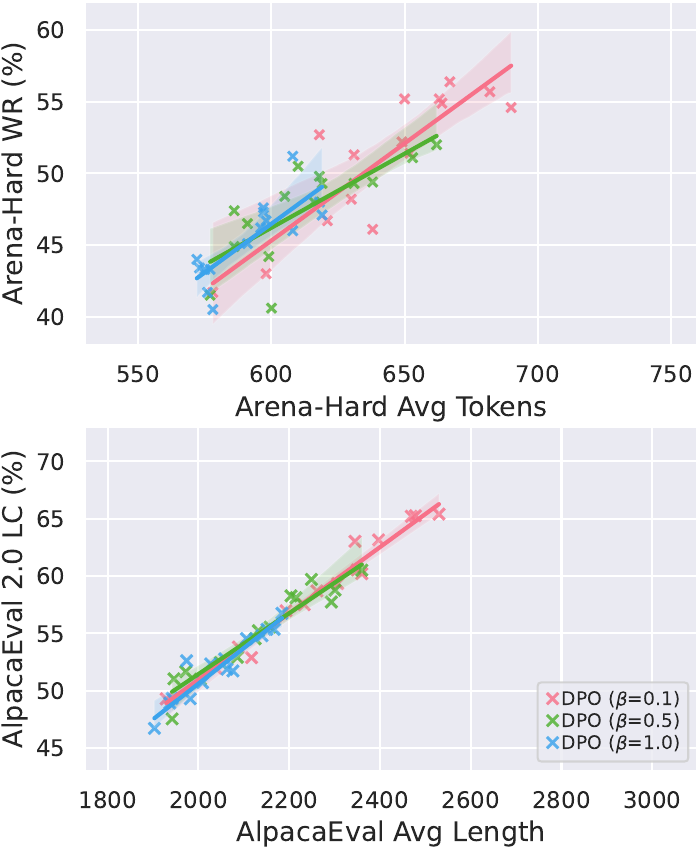}
        \caption{Hyperparameter search for DPO.}
        \label{fig:hyperparameter_tuning_dpo}
    \end{subfigure}
    \begin{subfigure}[t]{0.325\linewidth}
        \includegraphics[width=\linewidth]{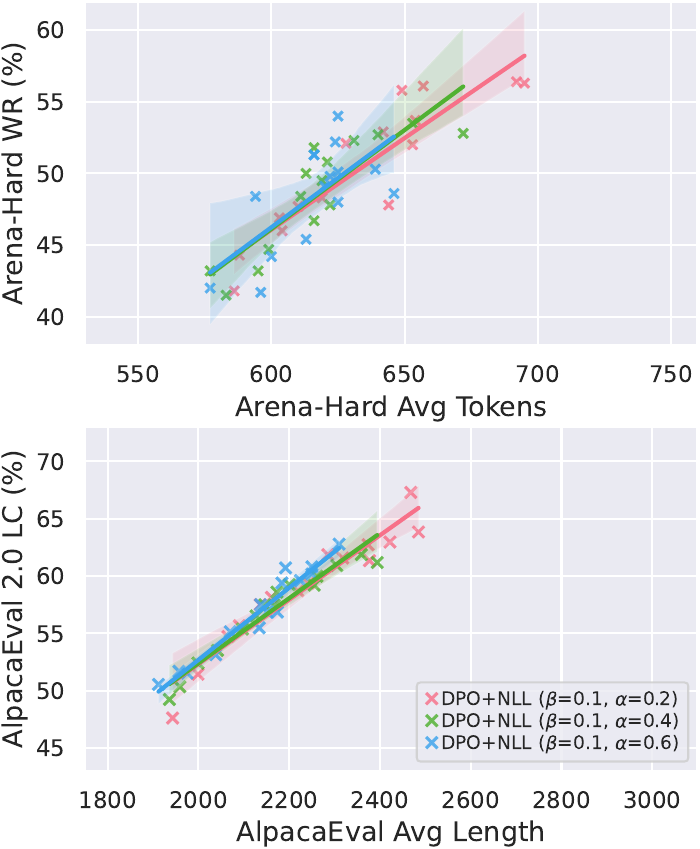}
        \caption{Hyperparameter search for DPO+NLL.}
        \label{fig:hyperparameter_tuning_dpo_nll}
    \end{subfigure}
    \begin{subfigure}[t]{0.325\linewidth}
        \includegraphics[width=\linewidth]{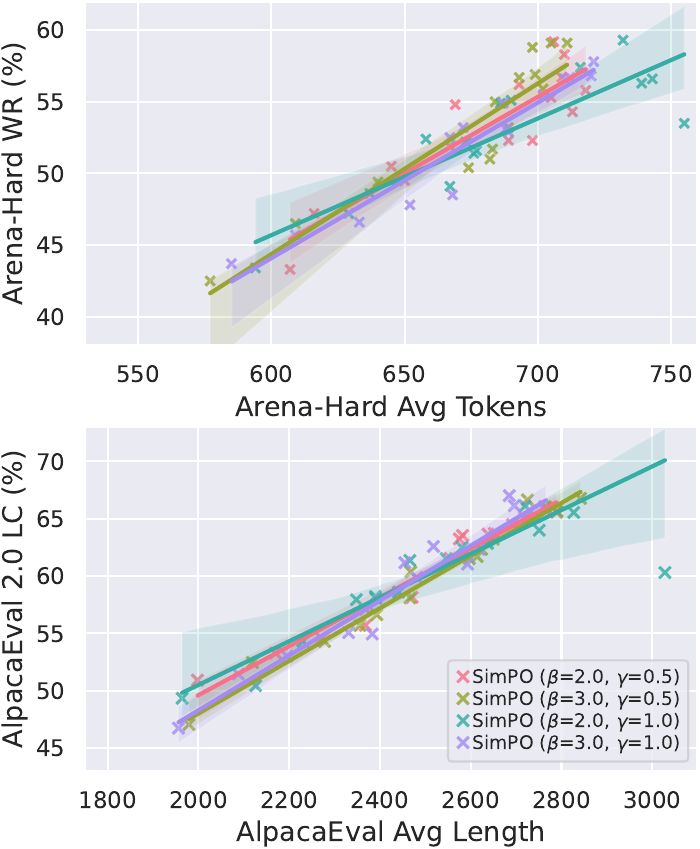}
        \caption{Hyperparameter search for SimPO.}
        \label{fig:hyperparameter_tuning_simpo}
    \end{subfigure}
    \begin{subfigure}[t]{0.325\linewidth}
        \includegraphics[width=\linewidth]{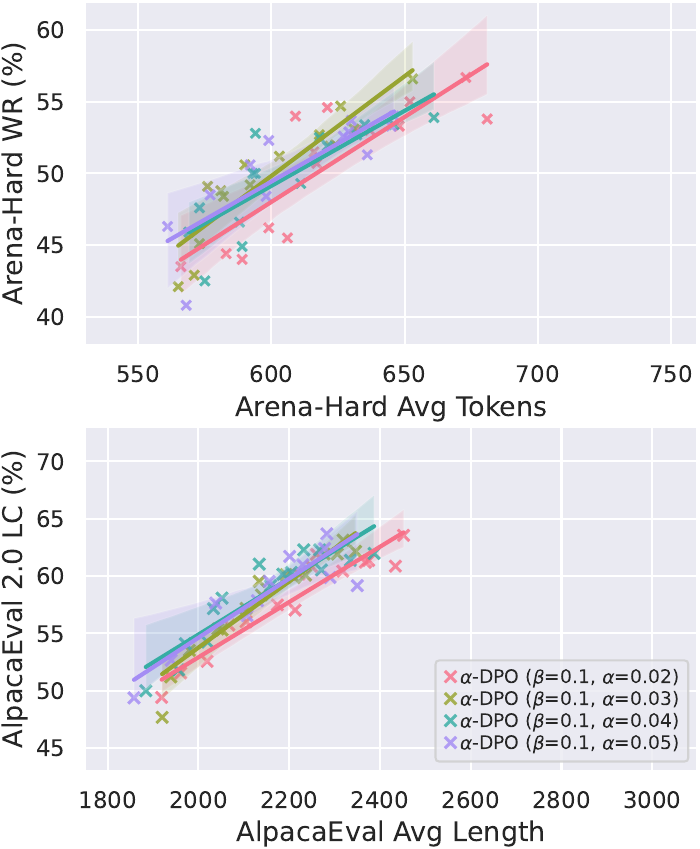}
        \caption{Hyperparameter search for $\alpha$-DPO.}
        \label{fig:hyperparameter_tuning_alpha_dpo}
    \end{subfigure}
    \begin{subfigure}[t]{0.325\linewidth}
        \includegraphics[width=\linewidth]{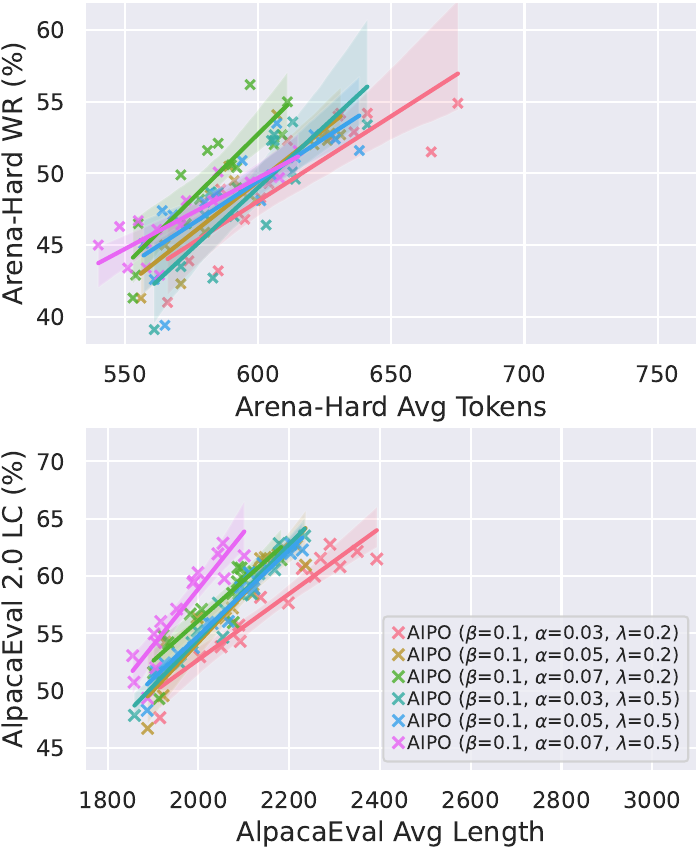}
        \caption{Hyperparameter search for AIPO.}
        \label{fig:hyperparameter_tuning_aipo}
    \end{subfigure}
    \caption{The detailed ablation of models and hyperparameters.}
    \label{fig:ablation_side_by_side}
\end{figure*}

Previous work suggests that the choice of hyperparameters is crucial for the training of non-iterative PO models~\cite{simpo}.
We also observe the same phenomenon in IPO training.
To ensure a fair comparison, we perform detailed hyperparameter tuning for the experiments presented in Sec.~\ref{subsec:optimize_dpo}.
We detailed the range of hyperparameter search for each method in Tab.~\ref{tab:hyperparameter}, and the results are shown in Fig.~\ref{fig:hyperparameter_tuning_dpo}-\ref{fig:hyperparameter_tuning_aipo}.
For each method, we select the best-performing model (located in the top left corner of the figure) to be included in the main paper.

\end{document}